\documentclass[pdflatex,sn-mathphys-num]{sn-jnl}

\usepackage{graphicx}%
\usepackage{multirow}%
\usepackage{amsmath,amssymb,amsfonts}%
\usepackage{amsthm}%
\usepackage{mathrsfs}%
\usepackage[title]{appendix}%
\usepackage{xcolor}%
\usepackage{textcomp}%
\usepackage{manyfoot}%
\usepackage{booktabs}%
\usepackage{tcolorbox}%
\usepackage{tabularx}
\tcbuselibrary{skins,breakable}%
\usepackage{enumitem}
\usepackage{subcaption}
\usepackage{algorithm}%
\usepackage{algorithmicx}%
\usepackage{algpseudocode}%
\usepackage{listings}%
\usepackage{multicol}

\theoremstyle{thmstyleone}%
\theoremstyle{thmstyletwo}%

\theoremstyle{thmstylethree}%
\newtheorem{definition}{Definition}%

\begin{document}

\title[Article Title]{Bridging visual saliency and large language models for  explainable deep learning in medical imaging}

\author*[1]{\fnm{Paul Valery} \sur{Nguezet}}\email{nguezetpaulvalery@gmail.com}

\author[1]{\fnm{Elie} \sur{Tagne Fute}}\email{eliefute@yahoo.fr}

\author[2]{\fnm{Yusuf} \sur{Brima}}\email{ybrima@uos.de}

\author[1]{\fnm{Benoit Martin} \sur{Azanguezet}}\email{azanguezet@gmail.com}

\author*[3,4]{\fnm{Marcellin} \sur{Atemkeng}}\email{m.atemkeng@gmail.com}

\affil[1]{\orgdiv{Department of Mathematics and Computer science}, \orgname{University of Dschang}, \orgaddress{\city{Dschang}, \postcode{67}, \country{Cameroon}}}

\affil[2]{\orgdiv{Computer Vision}, \orgname{ Institute 
of Cognitive Science}, \orgaddress{\street{ Osnabrück 
University}, \city{Osnabrueck}, \postcode{D-49090}, \state{Lower Saxony}, \country{Germany}}}

\affil[3]{\orgdiv{Department of Mathematics}, \orgname{Rhodes University}, \orgaddress{\street{Grahamstown}, \city{Eastern Cape}, \postcode{6140}, \state{Eastern 
Cape}, \country{South Africa}}}

\affil[4]{\orgdiv{National Institute for Theoretical and Computational Sciences (NITheCS)}, \orgaddress{\city{Stellenbosch}, \postcode{7600}, \country{Western Cape, South Africa}}}


\abstract{
The opaque nature of deep learning models remains a significant barrier to their clinical adoption in medical imaging. This paper presents a multimodal explainability framework that bridges the gap between convolutional neural network (CNN) predictions and clinically actionable insights for brain tumor classification, leveraging large language models (LLMs) to deliver human-interpretable diagnostic narratives. The proposed framework operates through three coupled stages. First, nine CNN architectures are extended with a dual-output hybrid formulation that simultaneously optimises a classification head and a segmentation head, enabling spatially richer feature learning. Second, visual saliency attribution methods, namely Grad-CAM, Grad-CAM++, and ScoreCAM, are applied to generate class-discriminative heatmaps, which are subsequently refined into binary tumor masks via an adaptive percentile thresholding pipeline. Third, the resulting masks are mapped onto the Harvard–Oxford cortical atlas to translate pixel-level evidence into named neuroanatomical structures, and the extracted findings are encoded into a structured JSON file that conditions three LLMs (Grok3, Mistral, and LLaMA) to generate coherent, radiological-style diagnostic reports. Evaluated on a dataset of 4,834 contrast-enhanced T1-weighted brain MRI images spanning three tumor classes, InceptionResNetV2 achieved the highest classification performance and Grad-CAM++ yielded the best segmentation overlap. Among the language models, Grok3 led in lexical diversity and coherence, while LLaMA achieved the highest readability score. By integrating visual, anatomical, and linguistic modalities into a unified pipeline, the framework produces explanations that are technically grounded and meaningfully interpretable, advancing the transparency and clinical accountability of artificial intelligence assisted brain tumor diagnosis.
}


%




\maketitle

\section{Introduction}\label{sec1}

Brain tumors represent a significant challenge in neuro-oncology, owing to their heterogeneity, variable malignancy, and significant impact on patient prognosis \cite{louis20162016}. Early and accurate diagnosis is critical for effective treatment planning, and magnetic resonance imaging (MRI) remains the modality of choice due to its high-resolution, non-invasive imaging capabilities \cite{bauer2013survey}. However, interpreting MRI scans, especially in complex tumor cases, can be time-consuming and prone to inter-observer variability, underscoring the need for automated and reliable decision-support systems \cite{el2014brain}.

To address these diagnostic challenges, deep learning (DL) methods, and convolutional neural networks (CNNs) in particular, have emerged as usefull  tools for automated medical image analysis \cite{litjens2017survey}. These computational models excel at automatically learning hierarchical feature representations from voluminous and multi-parametric MRI data, capturing intricate patterns that may elude conventional analysis \cite{alzubaidi2021review}. DL-based models have demonstrate considerable potential in performing critical tasks such as tumor detection, segmentation into distinct sub-regions (e.g., enhancing tumor, peritumoral edema, necrotic core) \cite{liu2023deep}, and non-invasive differential diagnosis and genomic characterization. These models offer a potential avenue for supporting radiological diagnoses. Despite their growing adoption, the opaque nature of these models often raises concerns about trust and clinical acceptance \cite{tjoa2020survey}. Medical professionals require not only accurate predictions but also transparency and justification for those predictions.

To address this, various explainability techniques, predominantly in the domain of visual saliency, have been introduced into diagnostic pipelines. 
Gradient-weighted Class Activation Mapping (Grad-CAM), its enhanced variant Grad-CAM++, and Score-weighted Class Activation Mapping (ScoreCAM)  generate spatial heatmaps that highlight the image regions most influential to the model's prediction, offering a visual explanation of the network's internal reasoning without modifying its architecture or training procedure \cite{selvaraju2017gradcam}. Although these methods have shown clinical utility in identifying diagnostically relevant regions, they carry important limitations. Saliency maps remain coarse, lack semantic grounding, and do not articulate specific radiological features (e.g., enhancement patterns, boundary characteristics, or anatomical location) that underpin a given prediction  \cite{guluwadi2024enhancing, brima2024saliency}. As a result, heatmaps alone are insufficient to bridge the gap between model output and actionable clinical reasoning.

Furthermore, bridging the translational gap between computational outputs and actionable clinical intelligence requires moving beyond purely visual explanations. The emerging integration of natural language generation (NLG), facilitated by large language models (LLMs), presents a perspective shift towards this goal \cite{valerio2025segmentation}. These models offer the potential to synthesize complex, multi-modal data; including quantitative imaging features, segmentation masks, and saliency maps into coherent, narrative-style reports articulated in clinically relevant terminology \cite{alsaad2024multimodal}. However, hybrid systems introduces significant challenges. Paramount among these is the imperative to ensure the factual accuracy and ontological consistency of generated text, preventing misrepresentation of model confidence \cite{mahaut2024factual}. Moreover, achieving robust cross-modal alignment where the generated narrative is precisely and reliably correlated with the visual evidence is non-trivial and necessitates sophisticated architectures capable of joint representation learning to maintain a verifiable link between image features and their descriptive linguistic counterparts.

In response to these limitations, this work introduces a novel multimodal explainability framework designed to set a linking bridge between opaque model predictions and clinically actionable insights. The proposed system is designed to integrate three distinct modalities tightly: computational visual saliency mapping, domain-aware anatomical contextualization, and natural language-based interpretation. The methodology is executed sequentially:
\begin{itemize}
    \item First, to strengthen the classification backbone, a dual-output hybrid architecture is applied across nine established CNN models, enabling each network to perform tumor-type classification and spatial mask estimation during training.
    \item Second, to move beyond classification and provide spatial localization of the model's focus, visual explanation techniques are employed. Grad-CAM and its enhanced variant Grad-CAM++, are utilized to generate coarse heatmaps. These heatmaps highlight the discriminative image regions primarily those corresponding to tumor tissue that most strongly influenced the model's classification decision, offering an initial layer of interpretability.
    \item Third, to refine this coarse localization into a precise morphological delineation, a semantic segmentation module processes the saliency output. The generated heatmap serves as an attention guide, enabling to segment and produce a binary mask. This mask isolates the tumor region while suppressing irrelevant background information. This step transitions from showing \textbf{where} the model looked to precisely defining \textbf{what} it identified as the pathological structure.
    \item Fourth, following the methodology discussed in \cite{valerio2025segmentation}, we overlay the mask onto the Harvard-Oxford cortical and subcortical structural atlas \cite{jenkinson2012fsl}. By performing this atlas-based parcellation, the framework can explicitly identify and report specific neuroanatomical structures infiltrated or displaced by the tumor. This  step translates pixel-based coordinates into anatomically meaningful labels, thereby providing a causally linked, domain-aware explanation for the model's prediction that aligns directly with clinical reasoning. The output of this step is  a structured JSON file based on the information extracted from the brain atlas mapping.
    \item Finally, the framework synthesizes these quantitative and spatial findings from the JSON file into a comprehensible clinical discourse using an LLM as a narrative engine. The LLM is conditioned on structured inputs comprising the classified tumor type, its probabilistic confidence, the specific anatomical structures involved (as defined by the atlas) and the salient imaging features inferred from the segmented volumes. This process generates a structured clinical narrative that summarizes the findings in a format akin to a radiological report. By fusing evidence from visual, anatomical, and linguistic domains, the framework delivers a holistic explanation that does not just indicate \textbf{where} the model looked but provides a semantically rich description of \textbf{what} it found and \textbf{why} it matters.
\end{itemize}
 This multimodal approach significantly improves the transparency and auditability of CNNs, thereby fostering greater trust and facilitating smoother integration into the clinical diagnostic pathway by aligning directly with the cognitive workflows and informational expectations of medical professionals.

The remainder of this paper is structured as follows. Section~\ref{sec:related_work} presents a review of prior work on explainability in medical imaging, including visual and textual approaches. Section~\ref{sec:methodology} details the proposed multimodal framework, including its classification, saliency, segmentation, anatomical mapping, and language generation components. Section~\ref{sec:results} reports quantitative and qualitative results across classification performance, visual explanation alignment, segmentation, and language quality. Finally, Section~\ref{sec:conclusion} concludes the paper and outlines future research directions.

\section{Related works}\label{sec:related_work}

Explainability has become a critical focus in the development of artificial intelligence (AI) systems for medical imaging, with researchers exploring both visual and textual methods to increase transparency and user trust. In particular, efforts have intensified around generating human-interpretable insights that align model predictions with clinically relevant cues. This section reviews prior research in three main domains: visual saliency methods, language model-based explanations, and integrated multimodal approaches that bridge the gap between image-based and language-based reasoning.

Visual saliency techniques have been widely adopted as a means of elucidating the internal decision-making processes of CNNs in medical applications. Brima and Atemkeng \cite{brima2024saliency} proposed a saliency-driven framework that leverages Grad-CAM to highlight predictive regions in brain tumor MRI scans, while quantitatively correlating these highlighted areas with statistical measures to validate clinical relevance. Their approach emphasized the importance of combining visual cues with quantitative validation to establish the reliability of AI-generated explanations. Similarly, Mahesh et al. \cite{mahesh2024} developed an explainability-enhanced classification pipeline based on EfficientNetB0, integrating Grad-CAM to visualize the most discriminative regions contributing to tumor predictions. Their results demonstrated that adding saliency maps not only improved model interpretability but also assisted practitioners in validating the AI’s focus during diagnosis. Despite their usefulness, these techniques typically lack deeper semantic understanding and remain limited to heatmap-based localization, which often provides ambiguous or anatomically imprecise explanations.

Language-based interpretability methods have emerged as a complementary avenue for enhancing model transparency. Sheng et al. \cite{wang2024interactive} introduced ChatCAD, an interactive diagnostic system that couples visual analysis with LLMs reasoning to facilitate user-guided medical interpretation. By prompting LLMs to respond to image-derived features, their framework enabled multi-turn conversations that dynamically adjust to clinical contexts. This work highlighted the potential of LLMs to synthesize complex visual patterns into natural language outputs that mirror expert-level descriptions. However, such approaches still depend heavily on the quality and alignment of visual input, and challenges remain in ensuring factual consistency and minimizing hallucinated content in generated reports.

Recent advances have explored deeper integration between visual saliency and language generation to enable multimodal explainability. Basu et al. \cite{basu2023radformer} introduced RadFormer, a transformer-based model incorporating both local and global attention to provide interpretable predictions in gallbladder cancer detection. In addition to achieving high accuracy, the model generated textual justifications aligned with relevant image regions, exemplifying the benefits of tightly coupled visual-textual reasoning. Similarly, Valerio et al. \cite{valerio2025segmentation} proposed a system that converts segmentation masks into descriptive radiology-style reports using LLMs. Their work demonstrated the feasibility of grounding textual explanations in anatomical segmentations, thereby increasing semantic clarity in model outputs. Also, recent work by Singh et al. \cite{singh2025unsupervised} introduced an unsupervised framework to enhance dementia detection interpretability by integrating CNN relevance maps with neuroanatomical features into a context-enriched explanation space, providing validated, multi-level post-hoc explanations that cater to the specific needs of both radiologists and neurologists. These multimodal strategies represent a promising direction, as they seek to contextualize predictions not only through highlighted image regions but also via narrative interpretations that can be directly communicated to clinicians.

Together, these contributions underscore a growing movement toward more human-centric and interpretable AI in medical imaging. Yet, despite significant progress, there remains a need for unified frameworks that link classification, visual saliency, saliency-derived segmentation, anatomical context, and clinically faithful textual reporting. It is important to note that the intent of the saliency-based segmentation mask in this work is not to compete with dedicated semantic segmentation algorithms such as UNet, which may achieve higher pixel-level segmentation accuracy when trained on annotated masks. Rather, segmenting directly from saliency heatmaps offers distinct and complementary advantages: it requires no additional  segmentation training data, making it particularly valuable in clinical settings where labelled masks are scarce or costly to obtain. Furthermore, saliency-derived segmentation is inherently explainable, as the resulting mask directly reflects the regions the classification model attended to when making its decision, preserving a transparent and traceable link between the model's prediction and the highlighted anatomy. It is also computationally lightweight, introducing no additional training overhead beyond the classifier itself. Together, these properties make saliency-based segmentation a practical and interpretable alternative in clinically constrained environments, where explainability and annotation efficiency are often prioritised over marginal gains in segmentation precision.

\section{Dataset and methods}\label{sec:methodology}
The proposed method has three main stages as summarised in Figure~\ref{Fig: architecture}. Stage 1 is the dual-head CNN training step, which takes the data as input and outputs the classification results and the saliency heatmap. These are then passed as input to Stage 2, which is the segmentation step that outputs the mask highlighting the tumor boundary. Stage 3 then takes place following the methodology in \cite{valerio2025segmentation}. However, the work in \cite{valerio2025segmentation} presents a methodology using the Julich Brain Atlas, a unique and comprehensive three-dimensional atlas, to output a medical report from a given mask as input. In contrast, we adopt this methodology using the Harvard-Oxford cortical atlas \cite{jenkinson2012fsl}, which is appropriate for 2D images and in line with our dataset. The stages are discussed in detail in the following subsections.
\subsection{Dataset}\label{sec:Dataset}

\begin{figure}
    \centering
    \includegraphics[width=0.7\linewidth]{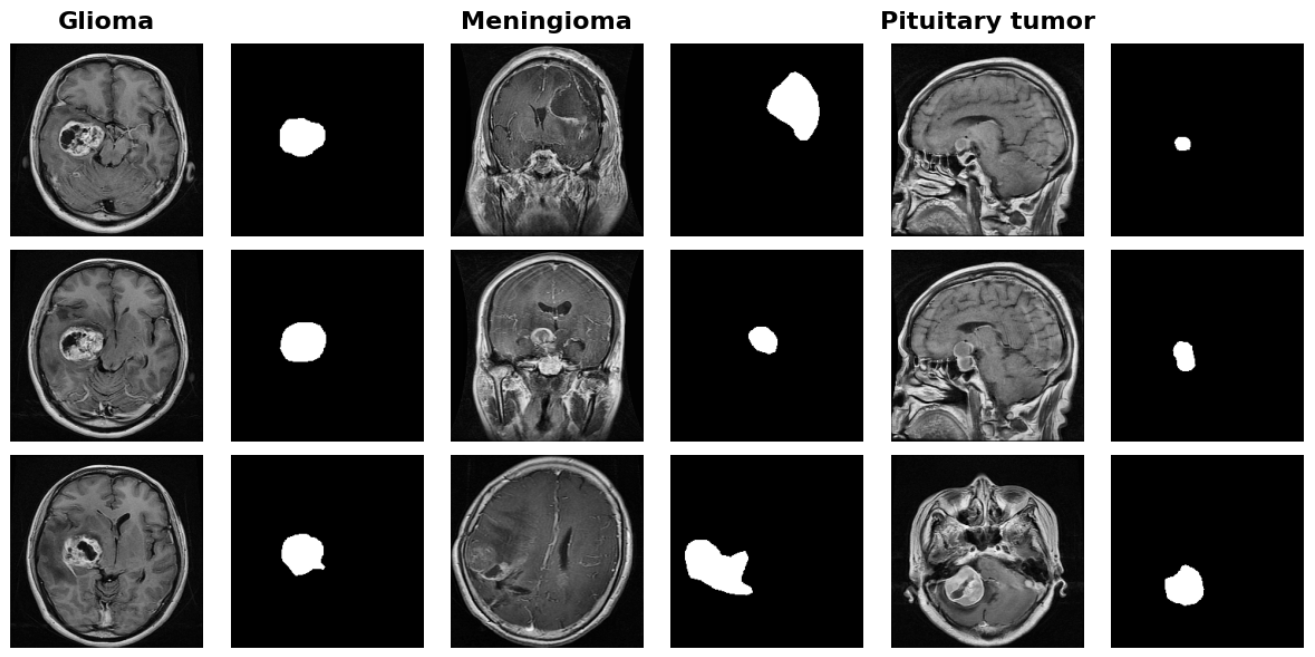}
    \caption{MRI scans of various brain tumors with tumor binary mask. This figure shows MRI images of the 3 brain tumor types studied, with their tumor binary mask localizing the tumor region.}
    \label{Fig: samples}
\end{figure}

\begin{figure}
    \centering
    \includegraphics[width=0.6\linewidth]{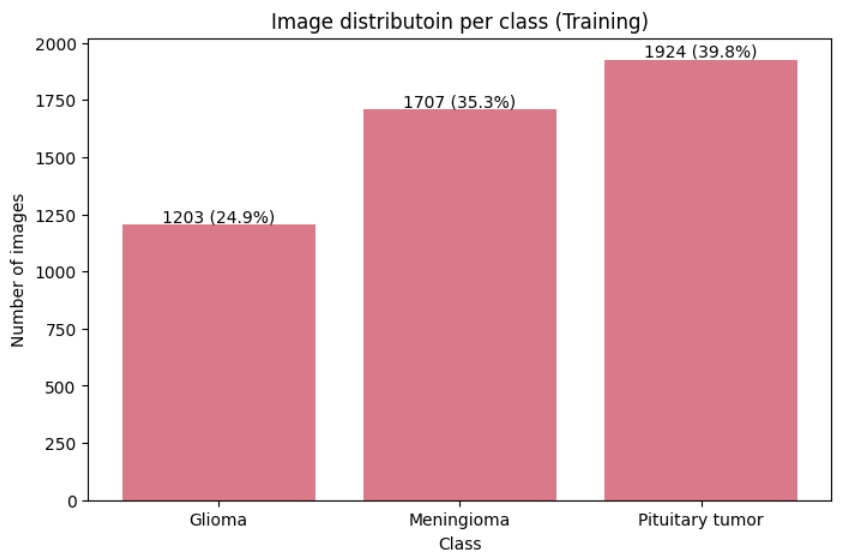}
    \caption{Bar chat showing brain MRI image distribution across tumor classes}
    \label{fig:sample_bar}
\end{figure}

The dataset used in this study is a curated and enhanced version of brain tumor MRI images derived from two publicly available datasets on the Kaggle platform 
. Both datasets contained brain MRI scans, each with the corresponding segmentation mask indicating the tumor region. The two datasets were merged into one after eliminating duplicate segmentation masks, which were identified through a visual inspection of the masks. The resulting dataset comprises 4,834 contrast-enhanced T1-weighted brain MRI images with paired binary segmentation masks, distributed across three tumor classes: 1,203 Gliomas, 1,707 Meningiomas, and 1,924 Pituitary tumors. Table~\ref{tab:datasets} summarizes the composition of each source dataset. Figure~\ref{Fig: samples} shows representative samples from each class alongside their corresponding tumor masks, and figure~\ref{fig:sample_bar}  illustrates the class distribution across the full dataset. 

All MRI images were subjected to a standardized preprocessing pipeline prior to model training. First, a contour-based cropping procedure was applied to each image and segmentation mask to eliminate dark background regions and focus the field of view on the anatomically relevant brain tissue. Both the cropped image and its mask were then resized to a uniform resolution of 225 × 225 pixels. Images were normalized to the range [0, 1] using min-max normalization. Masks were kept in their original 8-bit grayscale scale [0, 255] and subsequently binarized at a threshold of 127 prior to training.

\begin{table}
\centering
\caption{The 2 datasets modalities used to carry out experiments in this study}
\label{tab:datasets}
\renewcommand{\arraystretch}{1.2} 
\setlength{\tabcolsep}{4pt}
\begin{tabularx}{\linewidth}{l l c c l c}
\hline
\textbf{Source} & \textbf{Classes} & \textbf{Samples} & \textbf{Total} & \textbf{Modality} & \textbf{Segmented} \\
\hline
\multirow{3}{*}{Brain Tumor dataset 1 \cite{indrakumar_k_ravikumar_m_2025}} 
 & Meningioma      & 708   & \multirow{3}{*}{2192} & \multirow{3}{*}{MRI}   & \multirow{3}{*}{\checkmark} \\
 & Glioma          & 554  &                       &                         &                         \\
 & Pituitary tumor & 930   &                       &                         &                         \\
\hline
\multirow{3}{*}{Brain Tumor dataset 2 \cite{akter2024robust}}
 & Meningioma     & 999  & \multirow{3}{*}{2642} & \multirow{3}{*}{MRI} & \multirow{3}{*}{\checkmark} \\
 & Glioma       & 649 &                       &                        &                         \\
 & Pituitary tumor & 994  &                       &                        &                         \\
\hline
\end{tabularx}
\end{table}

\subsection{Dual-head CNNs training strategy}\label{sec:dual-head CNN}

Nine convolutional neural network architectures were employed as classification backbones: Densely Connected Convolutional Networks (DenseNet) \cite{huang2017densely}, Visual Geometric Group (VGG16 and VGG19 \cite{simonyan2014very}), DL with Depthwise Separable Convolutions (Xception) \cite{chollet2017xception}, Deep Residual Network (ResNet50, ResNet50V2) \cite{he2016deep}, a hybrid deep Inception and ResNet and EfficientNet \cite{tan2019efficientnet}, and Going deeper with convolutions (Inception) \cite{szegedy2015going}. These architectures were selected for their established performance in computer vision and medical image analysis tasks and to enable a comprehensive comparative evaluation across a diverse range of network designs. All models were implemented in TensorFlow and Keras, and initialized with ImageNet pre-trained weights to leverage transfer learning from large-scale visual representations

Rather than employing each backbone as a single-output classifier, we extend all nine architectures with a dual-output hybrid formulation, wherein a shared encoder extracts a common feature representation that simultaneously feeds two independent decoding heads: a classification head and a segmentation head. The classification head consists of a global average pooling layer followed by a fully connected layer with softmax activation, producing a probability distribution over the three tumor classes. The segmentation head is a lightweight decoder comprising four successive transposed convolutional layers with filter sizes of 256, 128, 64, and 32, each using ReLU activation, followed by a $1\times 1$ convolutional layer with sigmoid activation that produces a single-channel spatial probability map. A bilinear resizing layer is appended to enforce an output resolution of exactly $225\times 225$ pixels, matching the input image dimensions. This joint architecture enables each model to learn spatially richer and more discriminative representations during training, as the segmentation objective provides additional supervision over the spatial extent of pathological tissue, thereby strengthening the classification backbone that subsequently anchors the explainability pipeline.

To train the dual-output models, a combined loss function is minimised jointly over both heads. Let $y$ denote the ground-truth class label, $\hat{y}$ the predicted class probability vector, $M$ the ground-truth binary segmentation mask, and $\hat{M}$ the predicted mask. The total training loss $L_{\text{total}}$ is defined as:
\[
L_{\text{total}} = L_{\text{CE}}(\hat{y}, y) + \lambda L_{\text{Dice}}(\hat{M}, M),
\]
where $L_{\text{CE}}$ is the categorical cross-entropy loss over the classification head, $\lambda$ is a weighting coefficient balancing the two loss terms and $L_{\text{Dice}}$ is the  Dice Similarity Coefficient (DSC) loss on the segmentation head, defined as:
\[
L_{\text{Dice}} = 1 - \frac{2 |\hat{M} \cap M| + \epsilon}{|\hat{M}| + |M| + \epsilon},
\]
where $\epsilon$ is a small smoothing constant introduced for numerical stability.

All models were trained using the Adam optimizer with an initial learning rate of 0.0001, a batch size of 32, and a maximum of 30 epochs. To prevent overfitting, a learning rate reduction on plateau strategy was applied, halving the learning rate when no improvement in validation loss was observed for 5 consecutive epochs.

Training convergence was governed by a formal early stopping criterion derived from the joint validation behavior of both output heads. Specifically, training was stopped when the absolute difference between the validation loss and the predicted loss fell within a threshold defined by the standard deviation of the recent loss trajectory, expressed as:
\[
|L_\text{val} - L_\text{pred}| \leq \sigma(L),
\]
where \(L_\text{val}\) is the validation loss at the current epoch, \(L_\text{pred}\) is the predicted loss estimated from the recent loss trajectory, and \(\sigma (L)\) denotes the standard deviation of the loss values observed over the preceding patience window. When this condition is satisfied, training is considered to have converged, and the model weights yielding the lowest validation loss are retained. If the condition is not met, the combined loss \(L_\text{total}\) is recomputed and back-propagated to continue optimization. This criterion ensures that training terminates only when the model has reached a stable and reproducible optimum, rather than stopping prematurely due to transient fluctuations in the loss surface.

Training was conducted in a GPU-enabled environment. Classification performance was evaluated on the  test set using accuracy, precision, recall, and F1-score, reported both per class and as macro-averaged values. The DSC and Intersection over Union were  computed to assess the segmentation mask quality. 
The best-performing architecture, as determined by the macro-averaged F1-score on the test set, was selected to anchor the subsequent explainability pipeline.

\subsection{Visual saliency mapping}\label{sec:Visual Saliency Mapping}
A critical implementation consideration arising from the dual-head architecture concerns gradient isolation. Since the model produces two outputs simultaneously, a classification head and a segmentation head, a naive application of saliency methods would cause gradients from both branches to intermingle during backpropagation. Given that the segmentation branch is trained with a DSC loss weighted independently from the classification loss, its gradients carry spatial information about tumor boundary delineation rather than class-discriminative features. Allowing these gradients to contaminate the classification signal would produce heatmaps that reflect a mixture of segmentation and classification reasoning, undermining the interpretability goal. To prevent this, all saliency computations in this work are explicitly routed through the classification output layer only, ensuring that the resulting attribution maps faithfully represent the features that drove the tumor class prediction.

\begin{figure}
    \centering
    \includegraphics[width=1\linewidth]{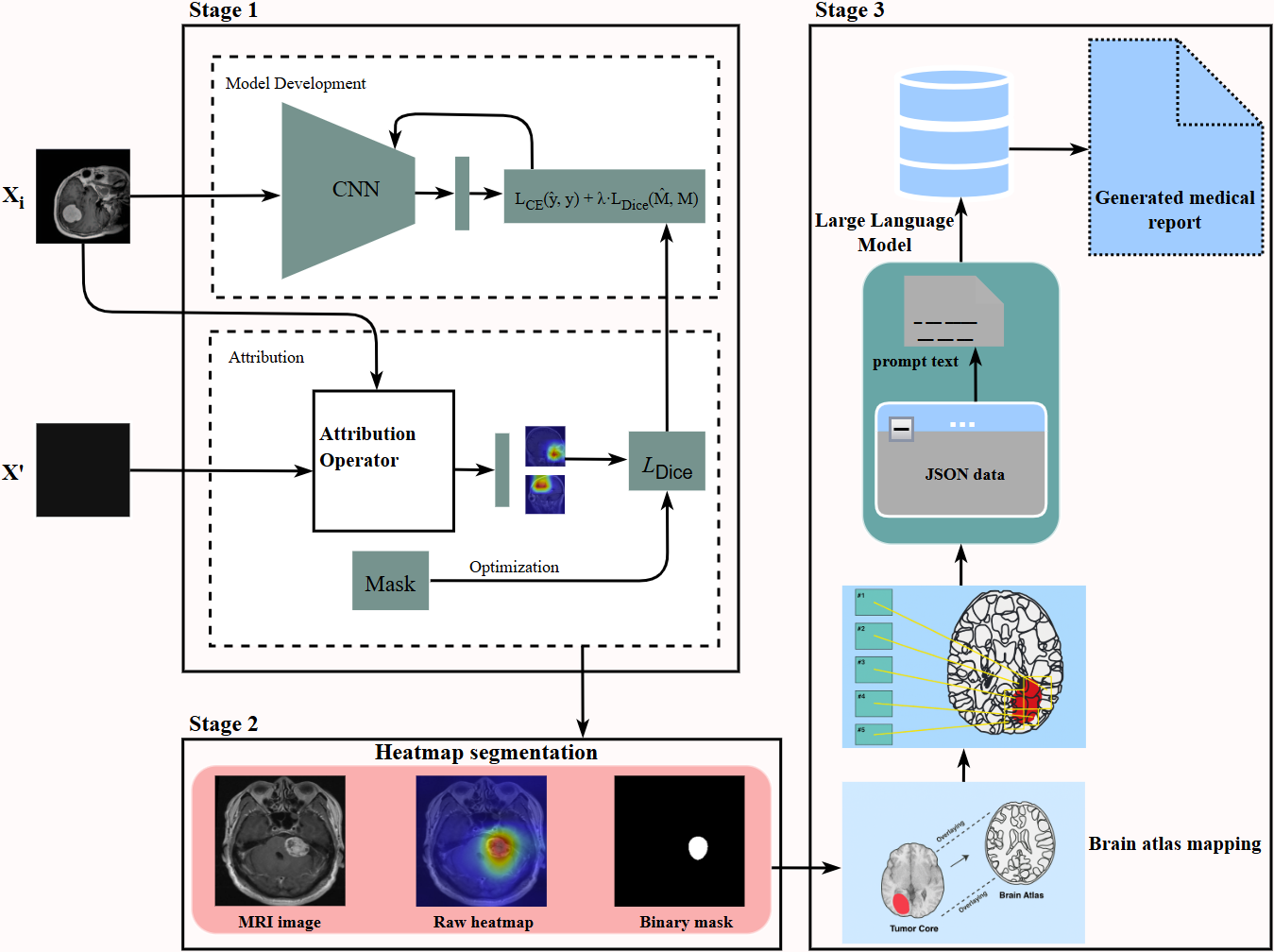}
    \caption{Pipeline of the proposed multimodal framework to CNN explainability. Stage 3 is adapted from \cite{valerio2025segmentation}.}
    \label{Fig: architecture}
\end{figure}

To analyze the attribution of the most salient features in medical images, these methods are integrated into the proposed framework as shown in Figure~\ref{Fig: architecture}. After the hybrid CNN model is trained with the corresponding tumor masks, it learns to map the features of the input image to their corresponding output labels and assimilate the location of the tumor through the masks. The goal is to understand what features the model prioritizes when making its predictions. To achieve this, a saliency attribution operator is applied to the trained model. This operator, which can be any one of the 3 saliency methods, highlights the most important or "salient" features within an input image that were critical to the model's decision-making process. This approach is fundamental for building trust in the model's inference mechanism, as it makes its reasoning transparent and understandable.

All three methods operate on the same target convolutional layer and are restricted to the classification output of the dual-head model. The resulting heatmaps are normalized to the range [0,1] and dynamically resized to match the spatial dimensions of each input image, ensuring precise spatial correspondence between the attribution map and the underlying MRI scan. Together, Grad-CAM, Grad-CAM++, and ScoreCAM provide a layered, mutually reinforcing basis for saliency attribution, combining gradient sensitivity and spatial precision for verification, to produce robust and clinically interpretable visual explanations.

\subsection{Heatmap segmentation}\label{sec:Segmentation}


To refine the coarse saliency maps into spatially precise tumor delineations, a post-processing pipeline was applied to convert each continuous heatmap into a binary mask. The pipeline consists of three sequential steps: percentile-based thresholding, removal of small connected components with an area below 50 pixels, and morphological closure with a disk structure element of radius r=3 to fill spatial gaps and produce coherent tumor regions. The complete procedure is described in Algorithm~\ref{alg:segment_heatmap}.

Rather than applying a fixed percentile threshold uniformly across all samples, a per-sample adaptive thresholding strategy was adopted to account for the substantial variability in tumor size and shape across the three classes. For each sample individually, the threshold percentile \(\alpha^*\) is selected by searching over \(\alpha \in [70, 97]\) and retaining the value that maximizes the DSC  between the thresholded heatmap mask and the ground-truth binary mask $M$:
\[
\alpha^{*} = \underset{\alpha}{\arg\max} \, \mathrm{DSC}\big(f_{\alpha}(H), M\big),
\]
where \(H\) is the normalized heatmap and \(f_\alpha(H) = \mathbf{1}[H \geq \text{percentile}(H, \alpha)]\) is the binary mask obtained at threshold \(\alpha\). This optimization is applied independently for each of the three saliency methods and for each test sample, yielding a sample-specific optimal threshold that adapts to the spatial footprint of the tumor present in that image.
The quality of the resulting segmentation masks is assessed by comparing them against the ground-truth tumor masks using two complementary metrics: the  DSC, which measures the overlap between predicted and reference regions, and the IoU, which penalizes over-segmentation more strongly. Both metrics range from 0 to 1, with 1 indicating perfect agreement with the ground truth.

\begin{definition}[Percentile]
Let $H \in [0,1]^{W \times H}$ be a heatmap with $N = W \cdot H$ intensity values sorted in non-decreasing order as $h_{(1)} \leq h_{(2)} \leq \cdots \leq h_{(N)}$. The $\alpha$-th percentile of $H$ is defined as:
\begin{equation}
    \mathrm{percentile}(H, \alpha) = h_{(\lceil \alpha N / 100 \rceil)}, \quad \alpha \in [0, 100]
\end{equation}
\end{definition}

\begin{definition}[Remove small objects]
Let $M \in \{0,1\}^{W \times H}$ be a binary mask and $s_{\min} \in \mathbb{R}^{+}$ a minimum area threshold. The operator $\mathrm{remove\_small\_objects}(M, s_{\min})$ returns a binary mask in which every connected foreground component $\mathcal{C}$ satisfying $|\mathcal{C}| < s_{\min}$ is suppressed:
\begin{equation}
    \mathrm{remove\_small\_objects}(M, s_{\min}) = M \setminus \bigcup \left\{ \mathcal{C} \subseteq M : |\mathcal{C}| < s_{\min} \right\}
\end{equation}
\end{definition}

\begin{definition}[Disk structuring element]
The disk structuring element of radius $r \in \mathbb{N}$ is the set of integer-coordinate pixels lying within Euclidean distance $r$ from the origin:
\begin{equation}
    \mathrm{disk}(r) = \left\{ (i, j) \in \mathbb{Z}^2 : \sqrt{i^2 + j^2} \leq r \right\}
\end{equation}
\end{definition}

\begin{definition}[Morphological closing]
Let $M \in \{0,1\}^{W \times H}$ be a binary mask and $S$ a structuring element. Morphological closing is defined as dilation followed by erosion:
\begin{equation}
    \mathrm{closing}(M, S) = (M \oplus S) \ominus S
\end{equation}
where $\oplus$ denotes morphological dilation and $\ominus$ denotes morphological erosion. This operation fills small holes and gaps within foreground regions while preserving overall shape.
\end{definition}

    \begin{algorithm}
        \caption{Thresholding and Morphological Filtering}
        \label{alg:segment_heatmap}
        \begin{algorithmic}[1]
            \Require Heatmap \( H \in [0,1]^{W \times H} \), ground-truth mask \(M \in \{0,1\}^{W \times H}\), search range \(\mathcal{A} = [70,97]\), minimum area \( s_{\text{min}} \in \mathbb{R}^+ \), closing radius \( r \in \mathbb{N} \)
            \Ensure Binary mask \( M \in \{0,1\}^{W \times H} \)
            
            \State \textbf{Flatten} all singleton dimensions of $H$ until it becomes a 2D array
            \For{each $\alpha \in \mathcal{A}$}
                \State $T \leftarrow \mathrm{percentile}(H, \alpha)$
                \State $M_{\alpha} \leftarrow \mathbf{1}[H \geq T]$
                \State $\mathrm{DSC}_{\alpha} \leftarrow \frac{2|M_{\alpha} \cap M| + \epsilon}{|M_{\alpha}| + |M| + \epsilon}$
            \EndFor
            \State $\alpha^* \leftarrow \arg\max_{\alpha \in \mathcal{A}} \; \mathrm{DSC}_{\alpha}$
            \State $T^* \leftarrow \mathrm{percentile}(H, \alpha^*)$
            \State $\hat{M} \leftarrow \mathbf{1}[H \geq T^*]$
            
            \State $\hat{M} \leftarrow \mathrm{bool}(\hat{M})$
            \State $\hat{M} \leftarrow \mathrm{remove\_small\_objects}(\hat{M},\, s_{\min})$
            \State $S \leftarrow \mathrm{disk}(r)$
            \State $\hat{M} \leftarrow \mathrm{closing}(\hat{M},\, S)$
            
            \State \Return $\hat{M}$
        \end{algorithmic}
    \end{algorithm}

\subsection{Brain atlas mapping}\label{sec:Brain Atlas Mapping}
To provide anatomical context for the detected tumor regions, the segmentation outputs were mapped to the Harvard–Oxford cortical atlas \cite{rushmore2022hoa2}. This step enables the transformation of visual explanations into clinically interpretable information by associating detected regions of interest (ROIs) with known brain structures.

First, ROIs are extracted from the binary segmentation mask obtained in the previous stage. As described in Algorithm~\ref{alg:extract_rois}, connected components are identified in the mask and characterized using their spatial coordinates, bounding boxes, and areas. Each component corresponds to a candidate tumor region detected from the heatmap segmentation.

To associate these regions with anatomical structures, the segmentation mask is aligned with the Harvard–Oxford atlas. Since the atlas and MRI images may differ in spatial resolution, the corresponding atlas slice is first resampled to match the dimensions of the segmentation mask. This ensures that each pixel in the mask corresponds to a valid anatomical label in the atlas.

\begin{definition}[Connected-component labelling]
Let $M \in \{0,1\}^{W \times H}$ be a binary mask. The function $\mathrm{label}(M)$ assigns a unique integer identifier $k \in \{1, 2, \ldots, K\}$ to each spatially connected foreground region $\mathcal{C}_k$, where connectivity is defined under 4-adjacency:
\begin{equation}
    \mathrm{label}(M) = L \in \mathbb{N}^{W \times H}, \quad L_{ij} = k \iff (i,j) \in \mathcal{C}_k
\end{equation}
with $L_{ij} = 0$ for all background pixels.
\end{definition}

\begin{definition}[Region properties]
Let $L \in \mathbb{N}^{W \times H}$ be a labelled mask with $K$ connected components. The function $\mathrm{regionprops}(L)$ computes, for each region $\mathcal{C}_k$, the following geometric descriptors:
\begin{itemize}
    \item \textbf{Coordinates}: $C_k = \{(i,j) : L_{ij} = k\}$, the set of pixel locations belonging to region $k$
    \item \textbf{Bounding box}: $B_k = (x_{\min}, y_{\min}, x_{\max}, y_{\max})$, the tightest axis-aligned rectangle enclosing $\mathcal{C}_k$
    \item \textbf{Area}: $A_k = |\mathcal{C}_k|$, the number of foreground pixels in region $k$
\end{itemize}
\end{definition}

    \begin{algorithm}[h]
        \caption{ROI Extraction from Binary Segmentation Mask}
        \label{alg:extract_rois}
        \begin{algorithmic}[1]
            \Require Binary mask \( M \in \{0,1\}^{W \times H} \), obtained from heatmap segmentation, representing tumor-relevant regions in an MRI image
            \Ensure List of region descriptors \( \mathcal{R} = \{(C_i, B_i, A_i)\}_{i=1}^n \), where \( C_i \) are pixel coordinates, \( B_i \) is the bounding box, and \( A_i \) is the area of the \( i \)-th region of interest (ROI)
            
            \State \textbf{Label} connected components: \( L \gets \text{label}(M) \)
            \State \textbf{Compute} region properties: \( \mathcal{P} \gets \text{regionprops}(L) \)
            \State \textbf{Initialize} ROI list: \( \mathcal{R} \gets \emptyset \)
            \State{\textbf{for} each region \( p \in \mathcal{P} \) \textbf{do}}
                \State \text{} \text{} \text{} \text{} \text{} Extract coordinates: \( C \gets p.\text{coords} \)
                \State \text{} \text{} \text{} \text{} \text{} Extract bounding box: \( B \gets (x_{\text{min}}, y_{\text{min}}, x_{\text{max}}, y_{\text{max}}) \)
                \State \text{} \text{} \text{} \text{} \text{} Extract area: \( A \gets p.\text{area} \)
                \State \text{} \text{} \text{} \text{} \text{} \textbf{Append} descriptor: \( \mathcal{R} \gets \mathcal{R} \cup \{(C, B, A)\} \)
            \State \textbf{end for}
            \State \Return \( \mathcal{R} \)
        \end{algorithmic}
    \end{algorithm}

Algorithm~\ref{alg:atlas_mapping} describes the atlas mapping procedure. For a given MRI slice, the corresponding axial slice from the atlas volume is extracted and resized to match the resolution of the ROI mask. The atlas labels are then queried at the pixel locations corresponding to the detected tumor regions. By counting the occurrence of atlas labels within the ROI mask, the method determines which anatomical regions are most affected. The output of this step is a table containing the detected brain regions, the number of voxels belonging to each region, and their percentage coverage within the ROI.

This atlas-based contextualization allows the framework to translate segmentation outputs into anatomically grounded descriptions (e.g., frontal cortex, temporal lobe), thereby improving the clinical interpretability of the explainability results.

\begin{definition}[Spatial resampling]
Let $S \in \mathbb{N}^{X \times Y}$ be a 2D atlas slice and $M \in \{0,1\}^{W \times H}$ a binary mask. The function $\mathrm{resize}(S,\, \mathrm{shape}(M),\, \mathrm{order}{=}0)$ resamples $S$ to the spatial dimensions $(W, H)$ of $M$ using nearest-neighbour interpolation (order $= 0$), which preserves discrete integer label values without introducing interpolation artefacts:
\begin{equation}
    S_r[i, j] = S\!\left[\left\lfloor i \cdot \frac{X}{W} \right\rfloor,\, \left\lfloor j \cdot \frac{Y}{H} \right\rfloor\right], \quad (i,j) \in \{0,\ldots,W{-}1\} \times \{0,\ldots,H{-}1\}
\end{equation}
\end{definition}

\begin{definition}[Nonzero indices]
Let $M \in \{0,1\}^{W \times H}$ be a binary mask. The function $\mathrm{nonzero}(M)$ returns the row and column index arrays of all foreground pixels:
\begin{equation}
    \mathrm{nonzero}(M) = \left( \{i : \exists\, j,\, M_{ij} \neq 0\},\; \{j : \exists\, i,\, M_{ij} \neq 0\} \right)
\end{equation}
\end{definition}

\begin{definition}[Value counts]
Let $L_i = \{l_1, l_2, \ldots, l_n\}$ be a sequence of atlas label values retrieved from foreground pixels. The function $\mathrm{value\_counts}(L_i)$ returns a frequency map $\mathcal{F}$ associating each unique label $l$ with its occurrence count:
\begin{equation}
    \mathrm{value\_counts}(L_i) = \mathcal{F}, \quad \mathcal{F}(l) = \left|\{k : l_k = l\}\right|, \quad \forall\, l \in \mathrm{unique}(L_i)
\end{equation}
\end{definition}

    \begin{algorithm}[h]
    \caption{Harvard-Oxford Cortical Atlas Mapping}
    \label{alg:atlas_mapping}
    \begin{algorithmic}[1]
    \Require Binary ROI mask $M \in \{0,1\}^{H \times W}$, atlas volume $A \in \mathbb{N}^{X \times Y \times Z}$, slice index $z$, label list $\mathcal{L}$
    \Ensure Table with \texttt{(region, voxel\_count, percentage)}
    
    \State Extract axial slice: $S \gets A[\colon,\colon, z]$
    \State Resample $S$ to match $M$: $S_r \gets \text{resize}(S, \text{shape}(M), \text{order}=0$
    \State Round and convert: $S_r \gets \text{round}(S_r).\text{astype}(\mathbb{N})$
    \State Find foreground indices: $(y_i, x_i) \gets \text{nonzero}(M)$
    \State Retrieve labels: $L_i \gets S_r[y_i, x_i]$
    \State Filter background: $L_i \gets L_i[L_i \neq 0]$
    \State Count occurrences: $C \gets \text{value\_counts}(L_i)$
    
    \State Construct DataFrame:
    \State \text{} \text{} \text{} \text{} [\textbf{Labels}:] $C.\text{index}$
    \State \text{} \text{} \text{} \text{} [\textbf{Counts}:] $C.\text{values}$
    \State \text{} \text{} \text{} \text{} [\textbf{Regions}:] $\mathcal{L}[C.\text{index}]$
    \State \text{} \text{} \text{} \text{} [\textbf{\% Coverage}:] $\text{(C.values/$\sum$ C.values)} \times 100$
    
    \State \Return Table sorted by $\texttt{voxel\_count}$ (descending)
    \end{algorithmic}
    \end{algorithm}

\subsection{JSON construction}\label{sec:JSON Construction}
In the next stage, the anatomical details derived from the atlas mapping are used to construct the JSON file. The structured JSON files helps to provide the LLMs  with precise, structured information, which will reduce the chances of hallucination by the LLMs, which often comes up as a major difficulty when language models generate text without truth base.

The JSON file contains clear and precise information including the model used for classification, the predicted class of the MRI image, the saliency method used to produce the heatmaps, the anatomical regions involved with the percentage of tumor occupation obtained from the atlas mapping, and the performance metrics for the heatmap segmentation.

\begin{figure}
    \centering
    \includegraphics[width=0.9\linewidth]{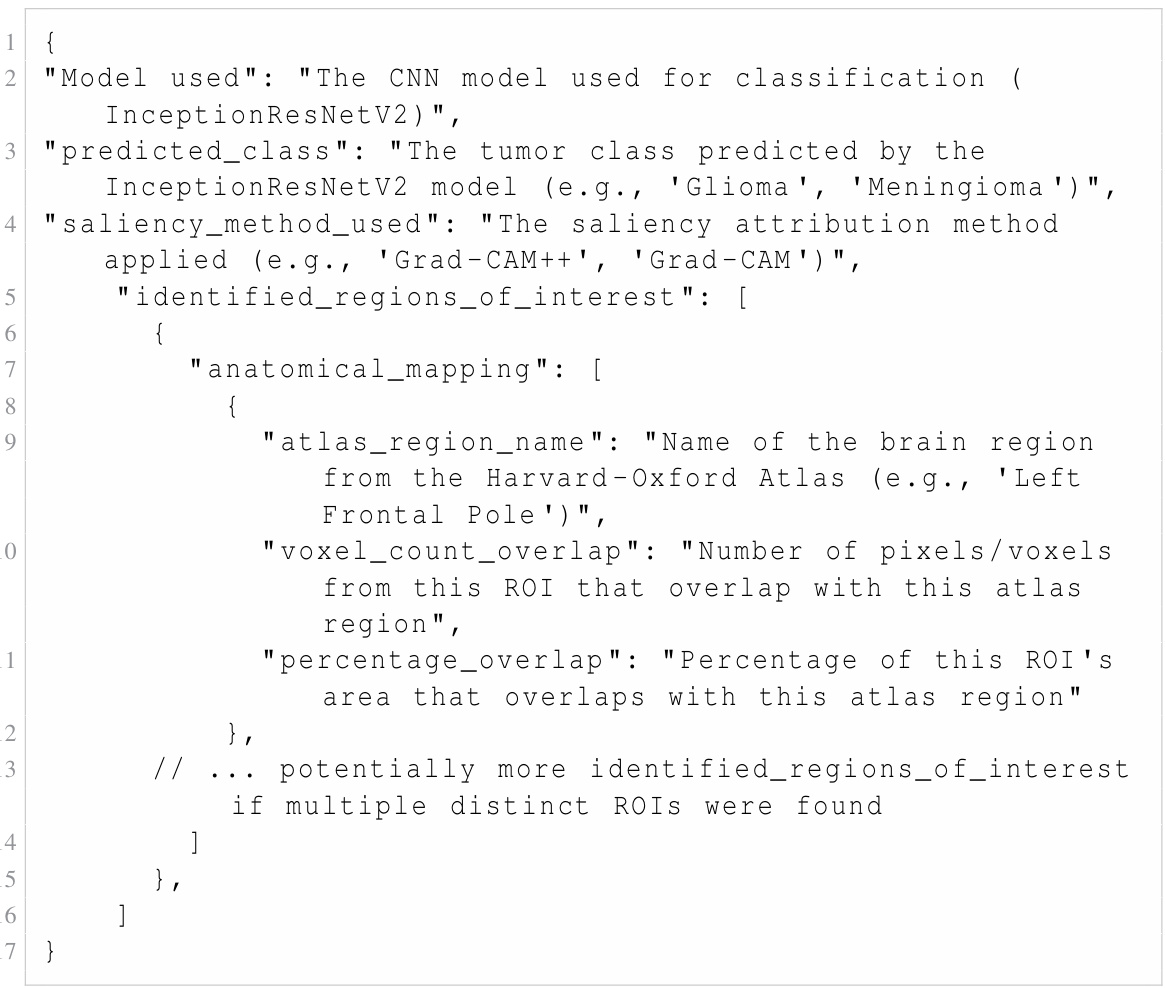}
    \caption{JSON file template }
    \label{Fig: JSON file}
\end{figure}

This representation provides precise and accurate information for the LLMs to generate valuable and trustworthy medical reports. The JSON used is shown in Figure~\ref{Fig: JSON file}.

\subsection{Language-based explanation generation}\label{sec:Language-based Explanation Generation}
The final stage of the framework sought to bridge the gap between computational outputs and clinical narratives. Here, we used three language models which will be based on the JSON file in order to generate valuable medical reports. The LLMs used include Grok3, Mistral, LLaMA which are open-accessed AIs available on the azure-openai platform.

\begin{figure}[h]
    \centering
    \includegraphics[width=1\linewidth]{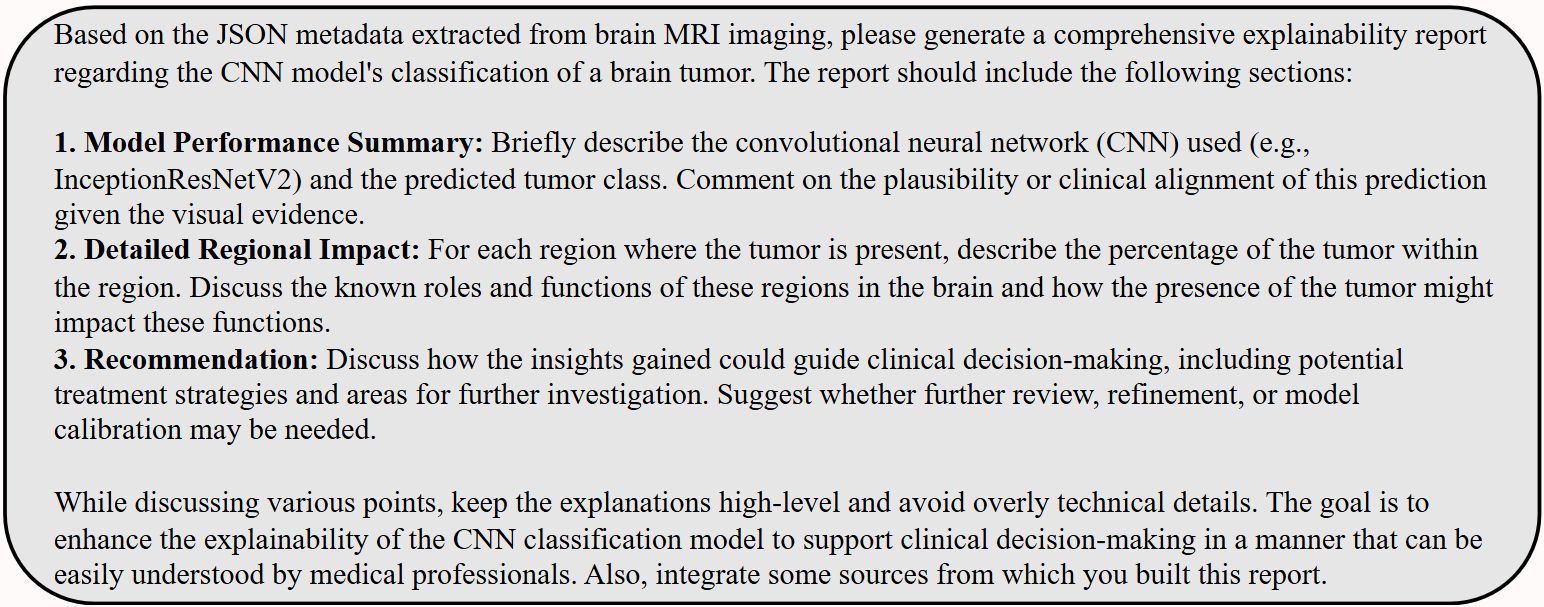}
    \caption{Prompt for LLM-based Explainability}
    \label{Fig:prompt}
\end{figure}

Following the structure of the JSON file, we then designed and elaborated a structured prompt to guide each LLM to generate coherent medical reports. In order to evaluate the quality of the generated reports, we assess the LLMs using quantitative metrics designed to capture essential linguistic qualities. These metrics include lexical diversity, readability, and coherence.

\textit{Lexical diversity} was quantified using two complementary measures. The Type-Token Ratio (TTR) was calculated as the ratio of unique words (types) to the total number of words (tokens) in a text, where a higher TTR indicates a greater variety of vocabulary \cite{jin2023guidelines}:
\[
\text{TTR} = \frac{\text{Number of Unique Words}}{\text{Total Number of Words}}.
\]
To account for the known sensitivity of TTR to the text length, Maas' Index was also employed. This logarithmic adaptation provides a more robust measure of lexical diversity, with lower values indicating greater diversity \cite{wang2024interactive}:
\[
\text{Maas} = \frac{\log(n_{\text{Tokens}}) - \log(n_{\text{Types}})}{\log(n_{\text{Tokens}})^2},
\]
where \(n_{\text{Tokens}}\) is the total number of words and \(n_{\text{Types}}\) is the number of unique words.

\textit{Readability} was assessed using the Flesch Reading Ease Score (FRES). This metric evaluates text comprehensibility based on average sentence length and average syllables per word. Higher scores indicate easier readability, with a score between 60–70 typically representing the standard for professional texts \cite{kaczmarek2022plenary}:
\[
\text{FRES} = 206.835 - (1.015 \times \text{ASL}) - (84.6 \times \text{ASW}),
\]
where \(\text{ASL}\) is the average number of words per sentence and \(\text{ASW}\) is the average number of syllables per word.

Finally, \textit{coherence} was measured using a Coherence Score (CohS) that quantifies the semantic flow between consecutive sentences. This is computed as the average cosine similarity of sentence embeddings for all adjacent sentence pairs in a report. Higher values indicate a stronger logical progression and better overall coherence \cite{bach2015pixel}:
\[
\text{CohS} = \frac{1}{N-1} \sum_{i=1}^{N-1} \cos(S_i, S_{i+1}),
\]
where \(N\) is the number of sentences, and \(S_i\) and \(S_{i+1}\) are the embeddings of consecutive sentences.

\section{Results}\label{sec:results}

\subsection{Brain tumor classification}\label{sec:Brain tumor Classification}

\begin{figure}[h]
    \centering
    \includegraphics[width=1\linewidth]{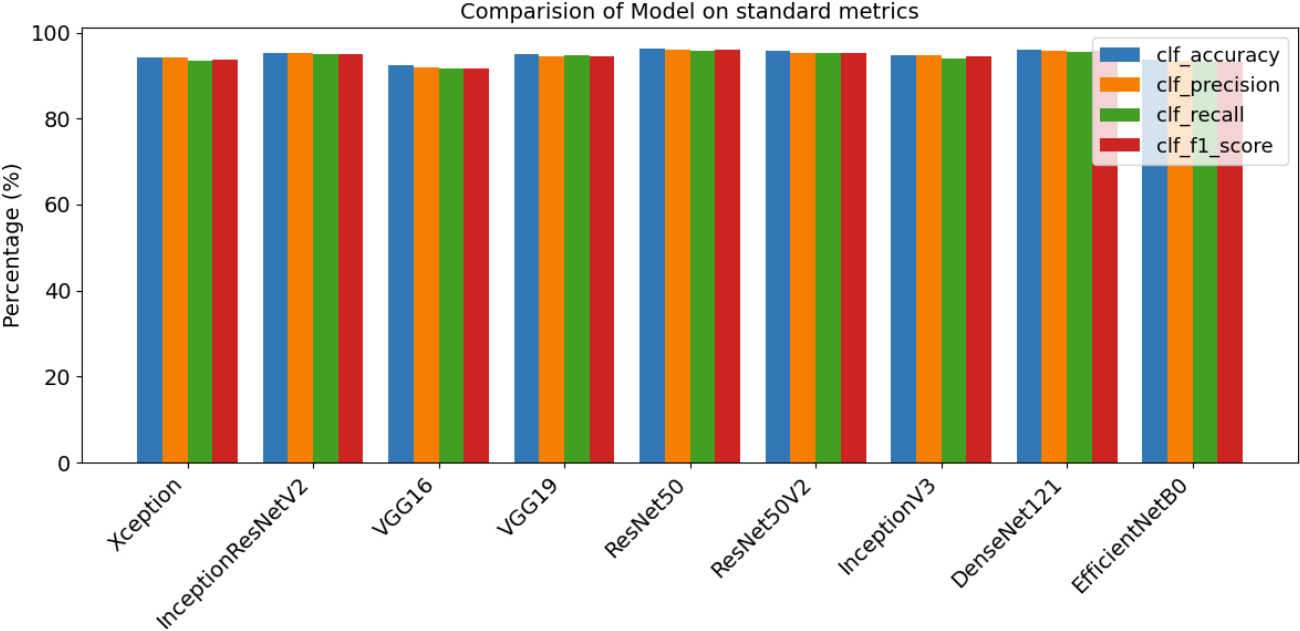}
    \caption{The evaluation metrics for each model are compared to assess their accuracy and robustness in brain tumor classification.}
    \label{fig:histogram}
\end{figure}

All nine CNN architectures were evaluated under the dual-output hybrid formulation described in Section~\ref{sec:dual-head CNN}, whereby each model simultaneously optimises a classification objective and a segmentation objective during training. Performance was assessed on the held-out test set using accuracy, precision, recall, and F1-score, and confusion matrix as illustrated in Figure~\ref{fig:histogram}  for the classification head. 

\begin{figure}[h]
    \centering
    \includegraphics[width=0.65\linewidth]{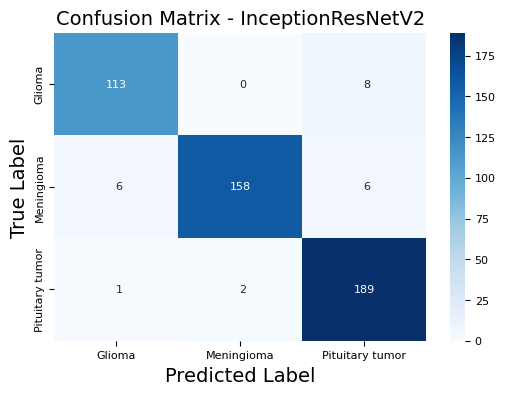}
    \caption{Confusion matrix for the top-performing model, InceptionResNetV2}
    \label{fig:conf_matrix}
\end{figure}

Among all evaluated architectures, the InceptionResNetV2 model achieved the highest overall performance across both output heads. For the classification task, InceptionResNetV2 attained an accuracy of 0.96 and achieved 0.95 across, precision, recall, and F1-score, which demonstrats consistent and balanced discrimination across all three tumor classes. Figure~\ref{fig:conf_matrix} shows the confusion matrix for InceptionResNetV2. This matrix shows that out of the 121 cases of glioma, 113 were correctly classified, and 8 cases were misclassified as pituitary tumor. Out of the 170 cases of meningioma, 158 were correctly classified, 6 cases were misclassified as glioma and 6 pituitary tumor. Out of the 192 cases of pituitary tumor, 189 were correctly classified,  2 cases were misclassified as meningioma and 1 as Glioma. 

\subsection{Saliency attribution}\label{sec:Saliency attribution}
Following the selection of InceptionResNetV2 as the best-performing backbone, Grad-CAM, Grad-CAM++, and ScoreCAM were applied to generate class-discriminative saliency maps for representative MRI samples drawn from each of the three tumor classes. These activation maps highlight the spatial regions most influential to the model's classification decision, providing a first layer of visual interpretability over the network's reasoning process.

\begin{figure}[h]
\centering
    \begin{subfigure}[b]{0.3\textwidth}
        \centering
        \includegraphics[width=0.35\linewidth]{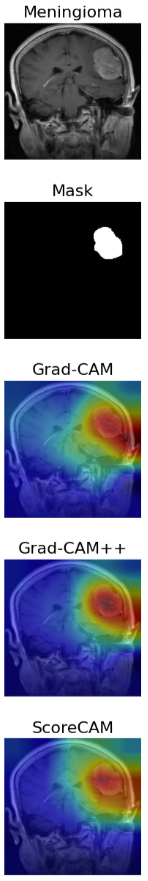}
        \label{fig:origin}
    \end{subfigure}
    \begin{subfigure}[b]{0.3\textwidth}
        \centering
        \includegraphics[width=0.35\linewidth]{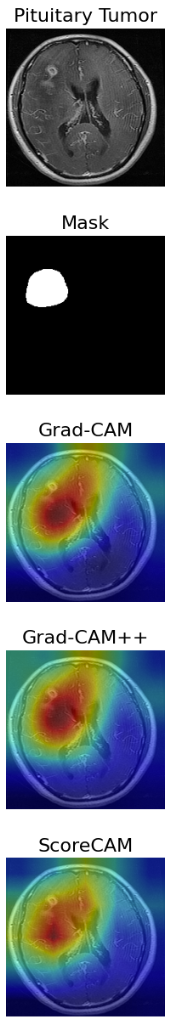}
        \label{fig:rawheatmap}
    \end{subfigure}
    \begin{subfigure}[b]{0.3\textwidth}
        \centering
        \includegraphics[width=0.35\linewidth]{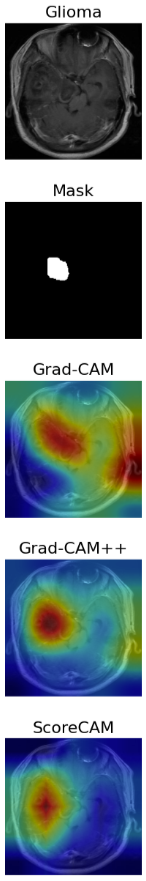}
        \label{fig:binarymask}
    \end{subfigure}
\caption{Saliency attribution of brain MRI images for each tumor type with Grad-CAM, Grad-CAM++, and ScoreCAM.}
\label{fig:saliency}
\end{figure}

Figure~\ref{fig:saliency} presents the saliency attribution results for all three types of tumors. For each sample, five rows are displayed: the original MRI image, the ground truth tumor mask, the Grad-CAM activation map, the Grad-CAM++ activation map, and the ScoreCAM activation map. Across all classes, all methods produce activation patterns concentrated in anatomically plausible regions, with highlighted areas broadly corresponding to the tumor tissue visible in the input scan. Grad-CAM++ consistently produces more focused and spatially precise activations compared to Grad-CAM and ScoreCAM, reflecting its per-neuron weighting scheme, which reduces diffuse background activation and sharpens localization around the most diagnostically relevant structures.


\begin{figure}
    \centering
    \includegraphics[width=0.8\linewidth]{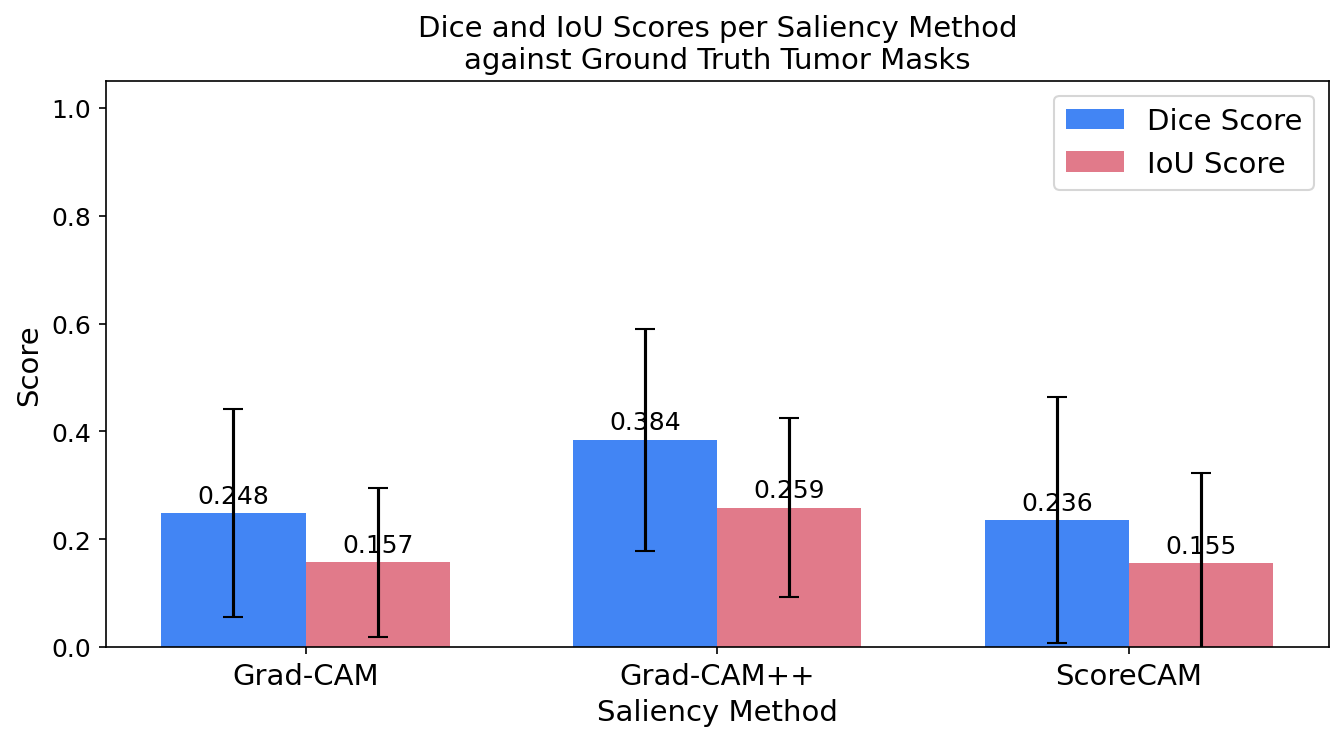}
    \caption{Comparing segmentation evaluation of the different saliency methods with DSC score and IoU.}
    \label{fig:dice_IoU}
\end{figure}

\subsection{Heatmap segmentation}\label{sec:Heatmap Segmentation}
To go beyond heatmap visualisation and towards spatially precise tumor delineation, the adaptive thresholding pipeline described in Section~\ref{sec:Segmentation}  was applied to the saliency maps generated by all three methods. For each sample, the optimal percentile threshold was determined independently by maximizing DSC against the ground-truth mask over the search range [70,97]. The optimal threshold varied substantially across samples, ranging from 71 to 96, depending on the spatial footprint of the tumor.

Figure~\ref{fig:dice_IoU} presents the DSC and IoU scores for all three saliency methods evaluated against ground-truth tumor masks. Grad-CAM++ achieved the highest segmentation performance with a mean DSC of 0.384 and IoU of 0.259, outperforming both Grad-CAM (DSC: 0.248, IoU: 0.157) and ScoreCAM (DSC: 0.236, IoU: 0.155). The superior localization of Grad-CAM++ is consistent with its per-location alpha weighting scheme, which concentrates activation on the most spatially specific regions rather than diffusing it across the broader feature map as Grad-CAM tends to do. ScoreCAM, despite being gradient-free and therefore immune to gradient saturation, produced slightly lower overlap scores.

\begin{figure}
\centering
    \begin{subfigure}[b]{0.8\textwidth}
        \centering
        \includegraphics[width=0.77\linewidth]{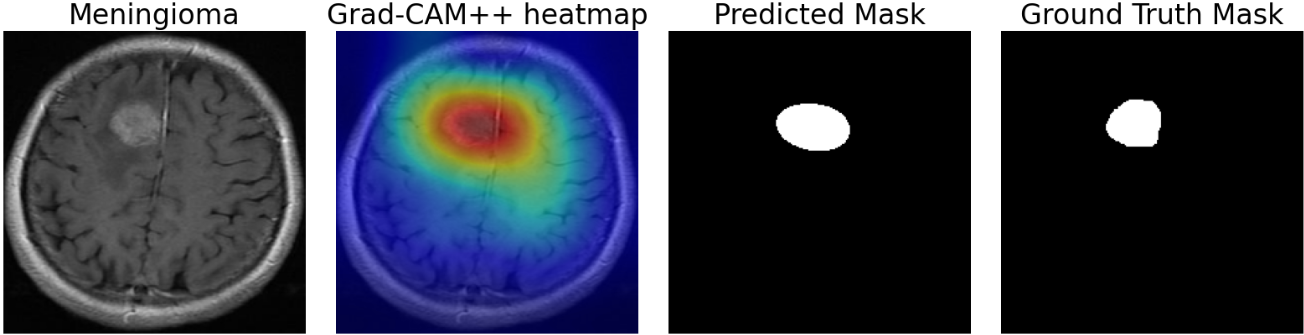}
        \label{fig:origin}
    \end{subfigure}
    \begin{subfigure}[b]{0.8\textwidth}
        \centering
        \includegraphics[width=0.77\linewidth]{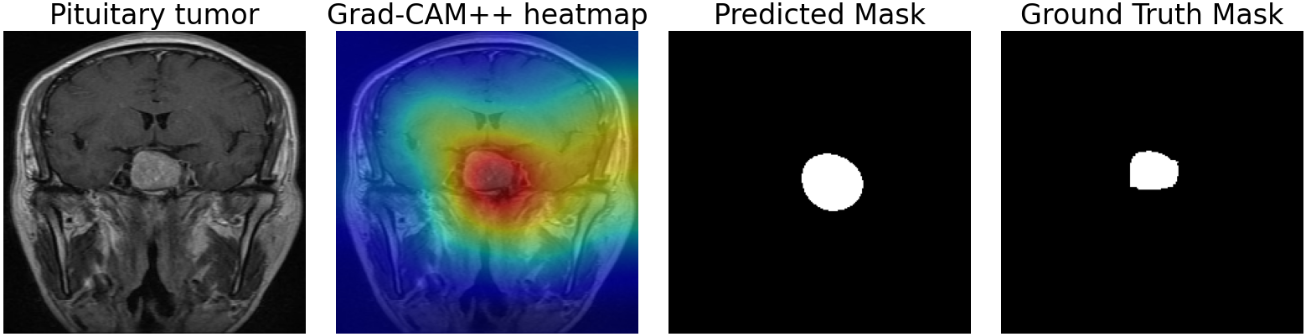}
        \label{fig:rawheatmap}
    \end{subfigure}
    \begin{subfigure}[b]{0.8\textwidth}
        \centering
        \includegraphics[width=0.77\linewidth]{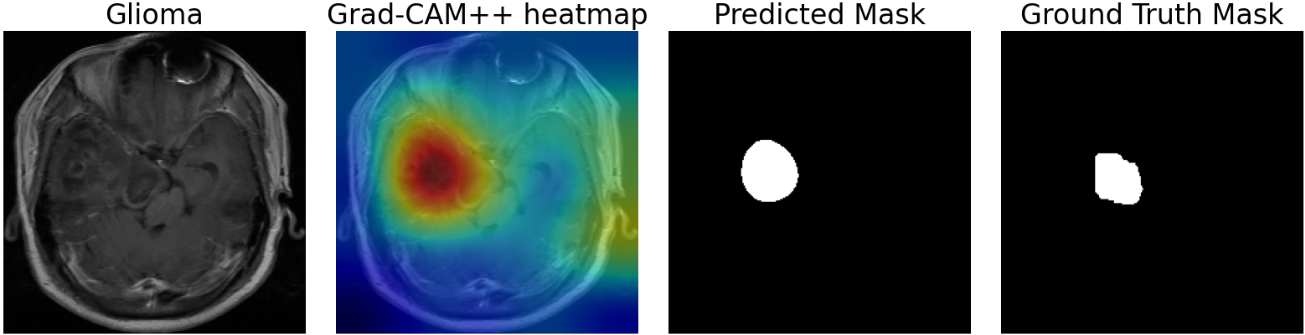}
        \label{fig:binarymask}
    \end{subfigure}
\caption{Heatmap segmentation for Meningioma, Pituitary tumor, and Glioma, respectively.}
\label{fig:segmentation}
\end{figure}

Figure~\ref{fig:segmentation} presents the segmentation visualization results for representative samples from all three tumor classes, showing the original MRI scan, the saliency heatmap, the predicted binary mask, and the ground-truth mask. Across classes, the predicted masks show qualitatively coherent localization, with the tumor region consistently captured within the predicted boundary. The remaining discrepancies between predicted and ground-truth masks are primarily attributable to the coarse spatial resolution of the saliency heatmaps relative to the precise pixel-level annotation of the ground-truth masks, a known limitation of class activation mapping methods.

\subsection{Textual explainability with LLMs}\label{sec:Textual Explainability}
\begin{figure}[h]
    \centering
    \includegraphics[width=0.7\linewidth]{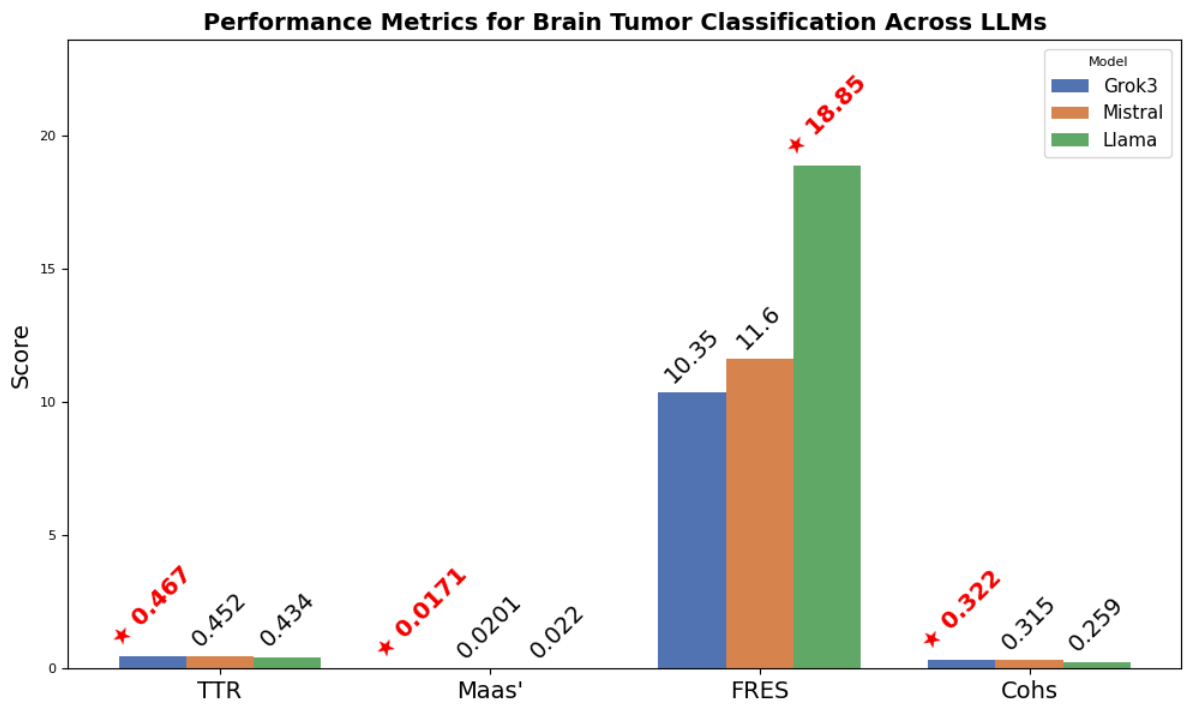}
    \caption{The performance metric for brain tumor classification across different LLMs.}
    \label{fig:llm-eval}
\end{figure}

The LLMs; Grok, Mistral, and LLaMA showed varying performance in generating medical reports explaining the model's classification and the saliency attribution. Figure~\ref{fig:llm-eval} highlights the different performance between models. Grok3 led in lexical diversity with the highest TTR of 0.467 and the lowest Maas' of 0.0171, indicating a medical report with a rich and diverse vocabulary. LLaMA outperformed in readability with an FRES of 18.85, while Grok showed a high performance in coherence. An example of the Grok3-generated medical report is shown in Box~\ref{box:medicalreport}.

\begin{tcolorbox}[
    enhanced, colback=gray!10, colframe=black!60, colbacktitle=gray!10, coltitle=black, fonttitle=\bfseries, fontupper=\small, title=Model Report: Brain Tumour Classification and Segmentation,
    label=box:medicalreport, width=\textwidth, boxrule=0.5pt, arc=3pt, top=6pt, bottom=6pt, left=6pt, right=6pt   
]

\noindent
\begin{minipage}[t]{0.49\textwidth}

\subsection*{Model Performance Summary}
The segmentation and classification of the brain tumor were performed using \textbf{InceptionResNetV2}, a convolutional neural network (CNN) known for its deep architecture and strong performance in image recognition. The model predicted the tumor class as \textbf{Meningioma}, a typically benign tumor arising from the meninges.

The prediction aligns with visual evidence from saliency maps generated via Grad-CAM++. Highlighted regions near the brain's surface and meninges are consistent with meningiomas, which grow along the dura mater and may compress adjacent brain structures. This supports the model’s classification and indicates effective identification of pathologically relevant areas.

\subsection*{Detailed Regional Impact}
The tumor affects multiple brain regions, with varying degrees of involvement quantified by percentage overlap of tumor voxels. Below are the affected regions, their functions, and potential clinical impacts:

\begin{itemize}
  \item \textbf{Insular Cortex (32.46\% overlap)}:  
    The insular cortex integrates emotional, sensory, and bodily state information, contributing to pain perception, self-awareness, and autonomic regulation. Tumor presence may cause disturbances in emotional processing, altered pain perception, or autonomic dysfunction (e.g., heart rate variability or gastrointestinal issues).
  \item \textbf{Cingulate Gyrus, Anterior Division (64.10\% overlap)}:  
    The anterior cingulate gyrus supports emotion regulation, decision-making, and pain processing. Significant tumor involvement may lead to difficulties in emotional regulation, impaired decision-making, or behavioral changes like impulsivity or apathy.
  \item \textbf{Cingulate Gyrus, Posterior Division (1.71\% overlap)}:  
    The posterior cingulate gyrus aids memory retrieval and spatial navigation. Minimal overlap suggests negligible impact, but subtle memory or navigation deficits may warrant monitoring.
  \item \textbf{Central Opercular Cortex (1.71\% overlap)}:  
    This region supports language processing and sensory integration. Minor overlap makes significant impact unlikely, but speech or sensory disruptions could occur with tumor progression.
\end{itemize}

The significant involvement of the anterior cingulate gyrus and insular cortex sug-
\end{minipage}
\hfill
\begin{minipage}[t]{0.49\textwidth}
gests potential pressure on areas critical for emotional and autonomic regulation, consistent with meningiomas’ compressive effects, possibly causing headaches, seizures, or cognitive changes.

\subsection*{Recommendation}
The segmentation analysis guides clinical decision-making by clarifying the tumor’s location and functional impacts. Recommendations include:

\begin{itemize}
  \item \textbf{Treatment Strategies}:  
    Surgical resection is a primary option for accessible meningiomas causing symptomatic compression. Stereotactic radiosurgery (e.g., Gamma Knife) may suit smaller tumors, while watchful waiting with regular imaging is appropriate for asymptomatic cases.
  \item \textbf{Symptom Management and Monitoring}:  
    Prioritize assessing emotional regulation, cognitive function, and autonomic symptoms via neuropsychological testing and monitoring for seizures or headaches.
  \item \textbf{Areas for Further Investigation}:  
    Contrast-enhanced MRI is recommended to confirm the meningioma diagnosis and assess edema or vascular involvement. Correlating model findings with clinical history and exams enhances diagnostic certainty. Biopsy or PET scans may rule out malignant variants if atypical features appear.
\end{itemize}

\subsection*{References}

    \begin{enumerate}
        \item Craig, A. D. (2009). How do you feel—now? The anterior insula and human awareness. \emph{Nature Reviews Neuroscience}, 10(1), 59-70.
        \item Bush, G., Luu, P., \& Posner, M. I. (2000). Cognitive and emotional influences in anterior cingulate cortex. \emph{Trends in Cognitive Sciences}, 4(6), 215-222.
        \item Leech, R., \& Sharp, D. J. (2014). The role of the posterior cingulate cortex in cognition and disease. \emph{Brain}, 137(1), 12-32.
        \item Goldbrunner, R., Minniti, G., Preusser, M., et al. (2016). EANO guidelines for the diagnosis and treatment of meningiomas. \emph{The Lancet Oncology}, 17(9), e383-e391.
    \end{enumerate}

In summary, the InceptionResNetV2 model’s segmentation results offer a roadmap for understanding the tumor’s impact and tailoring a multidisciplinary care approach, optimizing patient outcomes while minimizing risks to brain functions.

\end{minipage}

\end{tcolorbox}

\section{Conclusion}\label{sec:conclusion}

This paper discussed a multimodal explainability framework for brain tumor classification from MRI scans, designed to bridge the gap between opaque DL predictions and clinically actionable insights. The framework integrates four  coupled components: a dual-output hybrid CNN backbone for simultaneous classification and spatial mask estimation, visual saliency attribution, saliency-based segmentation, anatomical atlas mapping, and a structured LLMs reporting module. Together, these components provide a transparent, end-to-end pathway from raw MRI input to human-readable diagnostic narrative.

The atlas mapping step successfully translated pixel-level saliency evidence into named neuroanatomical structures using the Harvard–Oxford cortical atlas, providing an additional layer of clinical interpretability that purely visual explanations cannot offer. The three LLMs evaluated which are Grok3, Mistral, and LLaMA, demonstrated complementary strengths in generating structured diagnostic narratives from the JSON-encoded findings, with Grok3 leading in lexical diversity and coherence, and LLaMA achieving the highest readability score. Taken together, these results confirm that the proposed framework produces multimodal explanations that are technically sound and meaningfully interpretable for clinical practitioners.

Several limitations of the present study need to be acknowledged. The absence of prospective clinical validation means that the clinical utility of the generated reports has not yet been assessed by radiologists or neuro-oncologists in a real diagnostic workflow. The segmentation head, while effective as an auxiliary training signal, is not designed to replace dedicated segmentation architectures, and its spatial precision may be insufficient for surgical planning applications. Furthermore, while grounding LLM outputs in structured JSON representations reduces the risk of hallucination, occasional factual inconsistencies in generated narratives cannot be entirely excluded without expert clinical review.

Future work should extend this framework in several directions. Incorporating volumetric three-dimensional MRI data would enrich spatial context and improve the anatomical precision of both the segmentation outputs and the atlas mapping step. Critically, prospective validation in collaboration with clinical radiologists will be essential to assess the real-world usability and trustworthiness of the framework's outputs before any deployment in diagnostic settings. Finally, the development of domain-specific medical language models and adaptive prompting strategies represents a promising avenue for further reducing hallucination risk and improving the factual precision of generated narratives.

In summary, by integrating a strengthened hybrid classification backbone with adaptive saliency-driven segmentation, anatomical contextualisation, and language-based reporting, the proposed framework advances the transparency and clinical relevance of AI-assisted brain tumor diagnosis, contributing to the broader effort of making medical AI not only accurate but meaningfully accountable to the clinicians and patients it serves.

\subsection*{List of abbreviations}
MRI: Magnetic Resonance Imaging\\
CNNs: Convolutional Neural Networks\\
DL: Deep Learning \\
Grad-CAM: Gradient-weighted Class Activation Mapping\\ 
Grad-CAM++: Gradient-weighted Class Activation Mapping ++\\
ScoreCAM: Score-weighted Class Activation Mapping \\
NLG: Natural Language Generation \\
LLMs: Large Language Models\\
JSON: JavaScript Object Notation\\
AI: Artificial Intelligence \\
DenseNet: Densely Connected Convolutional Networks \\
VGG: Visual Geometric Group\\
Xception: Depthwise Separable Convolutions\\
ResNet: Deep Residual Network\\
Inception: Going deeper with convolutions\\
DSC: Dice Similarity Coefficient  \\
ROIs: Regions Of Interest

\subsection*{Acknowledgements}
Marcellin Atemkeng acknowledges the financial support of Rhodes University.
\subsection*{Author contributions}
P.V.N., M.A. and Y.B. designed and implemented the methodology. 
P.V.N. implemented the multimodal explainability framework, conducted the experiments, and wrote the main manuscript text.  E.T.F. , Y.B., B.M.A. and M.A. contributed to the visual saliency methodology and the related works section. E.T.F. and B.M.A. assisted with data preprocessing and the dual-head CNN training strategy. E.T.F. and M.A. supervised the work.  E.T.F. , Y.B., B.M.A. and M.A. proofread the manuscript. All authors reviewed and approved the final manuscript.

\subsection*{Funding}
This research was supported in part by the National Research Foundation of South Africa (Ref No. CSRP23040990793).

\subsection*{Data availability}
All datasets used and/or analysed during the current study are publicly available from their respective repositories through the references provided in the manuscript.

\section*{Declarations}
\subsection*{Ethics approval and consent to participate}
Not applicable.
\subsection*{Consent for publication}
Not applicable.
\subsection*{Competing interests}
The authors declare no competing interests.
\bibliography{sn-bibliography}

\end{document}